# *Open Set Domain Adaptation with Vision-language models via Gradient-aware Separation*

Haoyang Chen*

*Nanjing University of Information Science & Technology, Nan Jing, China*
*hychen2026@163.com*
**Corresponding author*

*Abstract:* Open-Set Domain Adaptation (OSDA) confronts the dual challenge of aligning known-class distributions across domains while identifying target-domain-specific unknown categories. Current approaches often fail to leverage semantic relationships between modalities and struggle with error accumulation in unknown sample detection. We propose to harness Contrastive Language-Image Pretraining (CLIP) to address these limitations through two key innovations: 1) Prompt-driven cross-domain alignment: Learnable textual prompts conditioned on domain discrepancy metrics dynamically adapt CLIP's text encoder, enabling semantic consistency between source and target domains without explicit unknown-class supervision. 2) Gradient-aware open-set separation: A gradient analysis module quantifies domain shift by comparing the L2-norm of gradients from the learned prompts, where known/unknown samples exhibit statistically distinct gradient behaviors. Evaluations on Office-Home show that our method consistently outperforms CLIP baseline and standard CoOp baseline. Ablation studies confirm the gradient norm's critical role.

*Keywords:* Open Set Domain Adaptation, CLIP, Gradient Norm, Vision-Language Models

## 1. Introduction

The rapid evolution of deep learning in image recognition, natural language processing and speech processing has enabled remarkable progress in cross-domain knowledge transfer, where unsupervised domain adaptation (UDA) techniques [1], [2] aim to bridge the distribution gap between labeled source domains and unlabeled target domains. Traditional closed-set UDA assumes identical label spaces across domains, a premise increasingly challenged by real-world scenarios where target domains inevitably contain novel categories unseen during training. This reality propels open-set domain adaptation (OSDA) [3], [4] into the research spotlight, demanding dual capabilities: (1) preserving discriminative knowledge of shared classes across domains while (2) robustly identifying unknown categories unique to the target domain.

  Current OSDA methodologies predominantly operate within single-modality paradigms, typically visual data, overlooking the rich semantic relationships inherent in multimodal representations [5], [6]. This limitation becomes particularly pronounced when dealing with domain shifts that alter feature correlations across modalities. For instance, in medical imaging, radiographic patterns (visual modality) and diagnostic reports (textual modality) exhibit interdependent shifts when adapting between hospital domains. The failure to model such cross-modal interactions leads to suboptimal alignment and error-prone unknown detection.

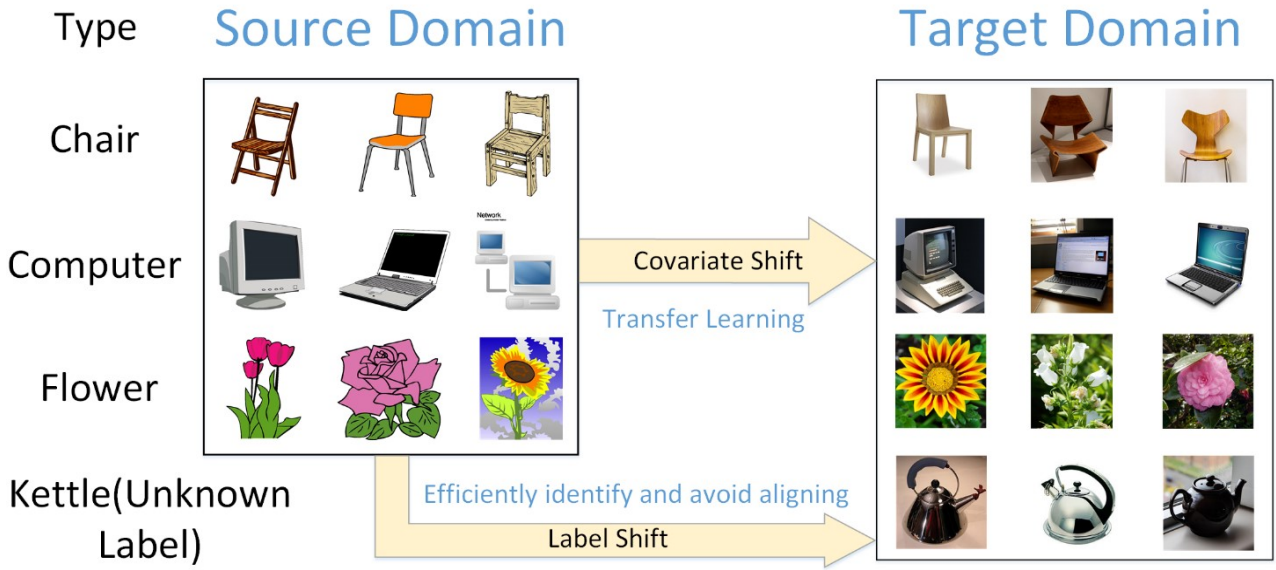

Figure 1: Illustration of covariate shift and open-set domain adaptation. Each row in the figure corresponds to a class in the source and target domains. For samples of the common classes that experience covariate shift, only the distribution of X changes. The bottom-most row in the target domain corresponds to unknown samples, for which models trained on the source domain should efficiently identify, and avoid aligning these samples with the source domain.

The emergence of multimodal models like CLIP (Contrastive Language-Image Pretraining) [7], [8] offers transformative potential. CLIP's joint image-text embedding space, pretrained on 400 million pairs, intrinsically links visual and textual semantics—a property that could disentangle domain-specific and category-specific variations. Yet, directly applying CLIP to OSDA faces critical barriers. First, static text prompts (e.g., "a photo of a {class}") fail to adapt to domain-specific contexts, such as medical versus natural imagery. Second, traditional unknown detectors relying on prediction entropy conflate hard-to-align known samples with true unknowns. Third, sequential alignment and detection stages propagate errors irreversibly; for example, misclassified unknowns contaminate feature alignment, which in turn degrades detection accuracy.

Existing OSDA methods fall into three categories, each with limitations. Adversarial frameworks like OSBP [2] use domain discriminators to separate shared and private label spaces but over-align visually similar known/unknown pairs. Prototype-based methods (e.g., OMEGA [3]) suffer from centroid drift under imperfect alignment and struggle with multimodal class distributions.

To address these challenges, we propose a unified framework that synergizes dynamic prompt learning with gradient-aware optimization. First, we introduce domain-conditioned prompt tuning, where learnable text prompts to mitigate domain shifts. These prompts dynamically adjust CLIP's text encoder, enabling semantic consistency across domains without unknown-class supervision. Second, we establish a gradient-norm-guided detection principle by showing that the gradient norm of the source and target domain samples can vary greatly.

We conducted extensive evaluations of our proposed method on benchmark dataset Office-Home, which are standard benchmarks for domain adaptation tasks. Our approach consistently outperformed existing single-modal OSDA methods and standard Cooperative Learning (CoOp) baselines across various adaptation scenarios. Notably, our method achieved superior performance in accurately classifying known classes while effectively identifying and managing unknown classes within the target domain. Ablation studies further substantiated the critical role of gradient norm analysis in enhancing the model's ability to separate known and unknown samples, underscoring the efficacy of our gradient-aware open-set separation strategy.

In summary, our approach not only addresses longstanding limitations in single-modal methods but also pioneers a new direction for leveraging pretrained vision-language models in open-set scenarios.

## 2. Related Works

**Out-of-distribution Detection**

Out-of-distribution (OOD) detection is a critical task in machine learning that aims to identify the difference of source domain, ensuring reliable performance in real-world applications. Recent years have seen significant progress in this field, with diverse methodologies emerging to address unique challenges. For instance, non-parametric distance-based methods like the K-nearest neighbors (KNN) approach proposed in "Out-of-Distribution Detection with Deep Nearest Neighbors" leverage deep feature embeddings without assuming Gaussian mixture distributions, achieving state-of-the-art performance on benchmarks [9], [10]. Gradient attribution analysis, exemplified by GAIA ,quantifies in-distribution (ID) and OOD differences through gradient attribution uncertainty, identifying anomalies like zero-inflation and channel-averaging to reduce false positive rates (FPR95)[11]. Feature and logit fusion approach such as ViM, combines feature space scores with class-specific logits by projecting features into orthogonal subspaces of training data principal components, introducing virtual logits to capture OOD-specific signals and outperforming traditional methods[12].Negative prompt learning addresses limitations in CLIP-based methods by learning category-specific negative prompts (e.g., "not a photo of [class]") to capture diverse OOD features, improving detection accuracy for semantically similar OOD samples[13]. These advancements reflect the increasing diversity and sophistication in the field, offering tailored solutions to challenges such as scalability, compatibility with contrastive learning frameworks [14], [15], and so on.

**Open Set Domain Adaptation**

Open Set Domain Adaptation (OSDA) [16], [17] solves the problems in realistic scenarios where target domain contains unknown categories which is absent in source domain compared with closed-set domain adaptation [18]. Pioneering work by Busto & Gall [19] introduced Assign-and-Transform-Iteratively (ATI), iteratively aligning shared classes via pseudo-labeling and linear transformations but relying on manual thresholds[20]. Saito et al. proposed Open Set Back-Propagation (OSBP)[21], leveraging adversarial training to dynamically separate unknowns via probability thresholds, but it is sensitive to the level of openness. To mitigate threshold dependency, CVPR 2019's Separate-to-Adapt (STA) [22] progressively disentangled known or unknown classes through weighted adversarial alignment. Recent researchers propose some innovations on the basis of them, such as Tian et al.'s moving-threshold estimation and gradual alignment. While these methods advance OSDA by balancing known-class alignment and unknown detection, challenges persist in dynamic openness adaptation and fine-grained semantic understanding of unknown categories.

**Vision-language Models**

Vision-language models have emerged as a transformative approach for understanding and reasoning across image and text modalities. Early works initially focused on designing dedicated architectures to align visual and linguistic representations, such as the dual-stream frameworks combining CNNs and RNNs (e.g. ViLBERT[23]). However, recent advancements have turned towards more unified frameworks that exploit large-scale pretraining to capture cross-modal semantics. Among these,

Contrastive Language-Image Pretraining (CLIP)[24], proposed by Radford et al. (2021), stands out for its simplification and effectiveness. CLIP introduces a novel contrastive learning approach that jointly trains a vision encoder (based on ViT or ResNet) and a text encoder (e.g. Transformer) on a massive dataset of image-text pairs which includes 400 million pairs. The core innovation lies in its training objective: instead of predicting pixel-level or region-level alignments, CLIP maximizes the similarity between global image embeddings and text embeddings via a contrastive loss. This formulation encourages the model to learn invariant representations that capture high-level semantics shared across modalities. By pretraining on the dataset created by Open AI, CLIP demonstrates remarkable zero-shot generalization to downstream tasks, surpassing supervised baselines on benchmarks like ImageNet and MSCOCO despite never fine-tuned by task-specific labels.

## 3. Methodology

### 3.1. Unknown Sample Recognition Algorithms

Common out-of-distribution (OOD) detection algorithms frequently utilize either features or the probability distribution of the model's final output for prediction.However, in this paper, the visual encoder of CLIP remains untuned, making it evidently infeasible to employ image features for OOD detection. Methods based on model outputs have been extensively studied in prior literature, particularly in [9], which proposes using the temperature-scaled softmax maximum as a detection metric—an approach we ultimately adopt during the testing phase.Nevertheless, when distributional shifts exist among in-distribution (ID) samples, it becomes apparent that the maximum softmax values of source domain samples often exceed 0.99, whereas those of target domain samples frequently fail to meet this threshold. Consequently, the decision boundary for identifying unknown samples cannot be reliably derived from the source domain data. Relying on manual threshold selection would inevitably introduce information leakage due to parameter tuning on the test set. Thus, we aim to explore alternative methods for OOD detection, seeking a metric that remains robust and does not exhibit significant variation with the distributional differences between source and target domain samples belonging to shared classes.

GradNorm [10] proposes to utilize the gradient norm of the model's parameters with respect to KL divergence as a metric to judge whether a test set sample is from unknown classes. Specifically, let p be the output of the model after softmax, and let $\boldsymbol{u} = [1/K, \ldots, 1/K]$ be uniform distribution, the KL divergence is defined as:

$$
\begin{aligned}
D_{\mathrm{KL}}(u||p) &= \sum_{i=1}^{K} u_i \log \frac{u_i}{p_i} \\
&= \sum_{i=1}^{K} \boldsymbol{u}_i \log \boldsymbol{u}_i - \boldsymbol{u}_i \log \boldsymbol{p}_i \\
&= \frac{1}{K} \times -\sum_{i=1}^{K} \log p_i - H(u)
\end{aligned}
\quad (1)
$$

Here, $H(u)$ denotes the entropy of u, which is a constant. GradNorm then computes the gradient of $D_{KL}(u||p)$ with respect to the final fully connected layer of the model and uses the L1 norm of this gradient as the metric. A larger norm indicates a higher likelihood that the sample is an in-distribution (ID) sample. Notice that we do not use $D_{KL}(p||u)$ here, as

$$
D_{\mathrm{KL}}(\boldsymbol{p}||\boldsymbol{u}) = -H(\boldsymbol{p}) + \frac{1}{K}\sum_{i=1}^{K} \log \frac{1}{\boldsymbol{p}_i}
\quad (2)
$$

degrades to entropy, where $D_{KL}(u||p)$ is similar to energy score [11] used in OOD detection.

However, this method cannot be directly applied to CLIP, because CLIP lacks a final fully connected layer [16]. A more critical issue is that experiments have revealed that the gradient norm of the loss for unknown samples with respect to the learnable prompts is larger than that for shared-class samples. This observation is exactly the opposite of the conclusion drawn by GradNorm. Therefore, it is necessary to thoroughly investigate how the gradient of the loss function with respect to the learnable prompts is calculated in the CoOp setting.

Let $f \in R^K$ denote the cosine similarity output by CLIP; let $v \in R^d$ represent the vector formed by concatenating the outputs of the learnable prompts after being processed by CLIP's text encoder, where $d = K \times 512$; let $z \in R^{512}$ denote the feature encoding of the image; let $w \in R^n$ denote the concatenated vector of learnable prompts. The gradient flow during SGD training process is illustrated as follows:

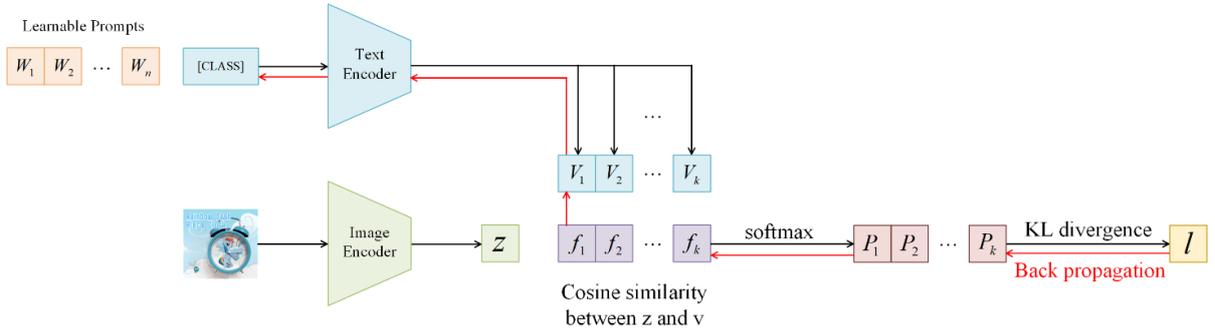

Figure 2: The gradient flow of learnable soft prompts. Black lines indicate forward process, and red lines indicate back propagation.

We wish to know $\frac{\partial l}{\partial w}$. By chain rule:

$$\frac{\partial l}{\partial w} = \left(\frac{\partial p}{\partial f}\frac{\partial f}{\partial v}\frac{\partial v}{\partial w}\right)^\top \frac{\partial l}{\partial p}$$
$$= \frac{\partial v^\top}{\partial w}\frac{\partial f^\top}{\partial v}\frac{\partial p^\top}{\partial f}\frac{\partial l}{\partial p} \quad (3)$$

where

$$\frac{\partial l}{\partial p} = \frac{1}{K}\frac{\partial}{\partial p}\left(-\sum_{i=1}^K \log p_i\right)$$
$$= \frac{1}{K}\left(-\frac{1}{p_1} - \frac{1}{p_2} \cdots -\frac{1}{p_K}\right)^\top \quad (4)$$

Notice that $p = \text{softmax}(f)$, therefore:

$$\frac{\partial p}{\partial f} = \begin{pmatrix} p_1 & 0 & \cdots & 0 \\ 0 & p_2 & \cdots & 0 \\ \vdots & \vdots & \ddots & \vdots \\ 0 & 0 & \cdots & p_K \end{pmatrix} - \begin{pmatrix} p_1 \\ p_2 \\ \vdots \\ p_K \end{pmatrix}(p_1 \quad p_2 \quad \cdots \quad p_K) \quad (5)$$

Since the cosine similarity f is computed via inner product of z and v,

$$\frac{\partial f}{\partial v} = \begin{pmatrix} \frac{\partial f_1}{\partial v} & \frac{\partial f_2}{\partial v} & \cdots & \frac{\partial f_K}{\partial v} \end{pmatrix} = \begin{pmatrix} z^\top & 0 & \cdots & 0 \\ 0 & z^\top & \cdots & 0 \\ \vdots & \vdots & \ddots & \vdots \\ 0 & 0 & \cdots & z^\top \end{pmatrix} \quad (6)$$

Therefore,

$$\frac{\partial l}{\partial w} = \frac{\partial v}{\partial w}^\top \frac{\partial f}{\partial v}^\top \frac{\partial p}{\partial f}^\top \frac{\partial l}{\partial p}$$

$$= \frac{1}{K} \frac{\partial v}{\partial w}^\top \begin{pmatrix} z & 0 & \cdots & 0 \\ 0 & z & \cdots & 0 \\ \vdots & \vdots & \ddots & \vdots \\ 0 & 0 & \cdots & z \end{pmatrix} \cdot \left( \begin{pmatrix} p_1 & 0 & \cdots & 0 \\ 0 & p_2 & \cdots & 0 \\ \vdots & \vdots & \ddots & \vdots \\ 0 & 0 & \cdots & p_K \end{pmatrix} - \begin{pmatrix} p_1 \\ p_2 \\ \vdots \\ p_K \end{pmatrix} \cdot \begin{pmatrix} p_1 & p_2 & \cdots & p_K \end{pmatrix} \right) \cdot \begin{pmatrix} -\frac{1}{p_1} \\ -\frac{1}{p_2} \\ \vdots \\ -\frac{1}{p_K} \end{pmatrix}$$

$$= \frac{\partial v}{\partial w}^\top \begin{pmatrix} z & 0 & \cdots & 0 \\ 0 & z & \cdots & 0 \\ \vdots & \vdots & \ddots & \vdots \\ 0 & 0 & \cdots & z \end{pmatrix} \begin{pmatrix} p_1 - \frac{1}{K} \\ p_2 - \frac{1}{K} \\ \vdots \\ p_K - \frac{1}{K} \end{pmatrix} \quad (7)$$

The primary difference between the above equation and GradNorm in gradient calculation lies in the fact that the latter's gradient lacks the initial term $\frac{\partial v}{\partial w}^\top$. As a result, the magnitude of GradNorm is nearly equivalent to the norm of $\begin{pmatrix} p_1 - \frac{1}{K} & p_2 - \frac{1}{K} & \cdots & p_K - \frac{1}{K} \end{pmatrix}^\top$, leading to higher scores for in-distribution (ID) samples. In CoOp, due to the presence of an additional constant matrix, the key factor determining the gradient norm's magnitude is the relationship between $N = \frac{\partial v}{\partial w}^\top$ and $M = \begin{pmatrix} z & 0 & \cdots & 0 \\ 0 & z & \cdots & 0 \\ \vdots & \vdots & \ddots & \vdots \\ 0 & 0 & \cdots & z \end{pmatrix} \begin{pmatrix} p_1 - \frac{1}{K} \\ p_2 - \frac{1}{K} \\ \vdots \\ p_K - \frac{1}{K} \end{pmatrix}$.

In fact, after temperature scaling and softmax, the cosine similarity of ID samples often results in one class dominating with a probability of 0.99, while the other classes combined account for only 0.01. In such cases, only the column corresponding to the dominant class (0.99) in matrix M exhibits significant values, while the remaining columns are nearly zero. When multiplied by N, only a small fraction of values deviate noticeably from zero, leading to smaller gradients. On the other hand, the probability distribution of unknown samples often contains multiple non-zero values—for example, three values of 0.3 each—causing more columns in M to have non-zero values. When multiplied by N, this trend is further amplified.

In summary, under the settings of this paper, samples with larger gradient norms should be considered unknown, and vice versa. By selecting a threshold based on source domain samples, we ensure that a high proportion (90%) of source domain samples are identified as ID. Experiments demonstrate that the threshold obtained this way maintains strong discriminative ability in the target domain, whereas the performance of the maximum softmax score degrades significantly.

## 3.2. Uncertainty-aware Pseudolabel Generation

To address the challenge of open-set domain adaptation where target domains contain both shared and unknown classes, we propose to explicitly handle these two types of samples differently. The key insight stems from the observation that forcing unknown-class samples into existing source categories amplifies error propagation, while treating all target samples as shared classes limits model discriminability. Our method first separates target samples into shared/unknown categories through gradient norms mentioned above. For identified shared-class samples, we enforce prediction consistency with their pseudo-labels through cross-entropy minimization to maintain class discriminative power. For uncertain samples likely belonging to novel categories, we instead regularize their output distributions to approximate a uniform distribution via KL divergence minimization, preventing overconfident predictions while encouraging representation openness. This bifurcated approach allows simultaneous preservation of source knowledge and adaptive discovery of target-specific novelty, effectively balancing closed-set alignment and open-set rejection without requiring prior knowledge of unknown class quantities. The final loss can be formulated as:

$$\mathcal{L} = \mathcal{L}_{\text{ce}}(x^s, y^s) + \alpha \boldsymbol{p}_{\hat{y}} \cdot \mathcal{L}_{\text{ce}}(\boldsymbol{x}^t, \hat{y}) + \beta D_{\text{KL}} \tag{8}$$

## 4. Experiments

### 4.1. Datasets

To test the performance of our proposed model, we conduct comprehensive experiment on public Domain Adaptation dataset named Office Home. The Office-Home dataset serves as a benchmark for evaluating Transfer Learning algorithms in domain adaptation tasks. It consists of 15,500 images across four distinct domains, each containing 65 categories with approximately 70 images per category. The domains are:

1. Art: Comprising 2,427 images of sketches, paintings, and decorative objects, characterized by abstract styles and artistic renditions.
2. Clipart: Featuring 4,365 simplified exaggerated illustrations with solid color backgrounds.
3. Product: Consisting of 4,439 object images without backgrounds, which is a large variety of e-commerce product shots.
4. Real-World: Including 4.357 photographs captured in natural setting diverse real-world object appearances.

The dataset provides labeled annotations for each image, enabling the model to train and evaluate. By emphasizing domain shifts between different domains, Office-Home challenges the capability of model's generalization across diverse data distributions. Researchers often utilize this dataset to develop robust methods which handle cross-domain variations.

### Metrics

**CCR@FPR10 (Correct Classification Rate at FPR=10%)**

CCR@FPR10 jointly assesses the ID classification accuracy of samples that are identified ID by the model. Specifically, it measures the proportion of ID samples that are not only correctly identified as ID, but also accurately classified into their true ID category when the decision threshold is set to achieve a 10% FPR on OOD samples.

This metric addresses the limitation of pure detection metrics by ensuring models not only reject OOD samples but also maintain ID classification performance at a low FPR. Higher CCR@FPR10 values indicate better classification performance of the model.

**FPR95 (False Positive Rate at 95% True Positive Rate)**[10]

FPR95 measures the FPR (percentage of OOD samples misclassified as ID) when the TPR (percentage of ID samples correctly identified) is fixed at 95%. The metric reflects model performance under high recall rate, where maintaining ID detection sensitivity is prioritized. Lower FPR95 values (closer to 0) denote better OOD detection robustness.

**AUROC (Area Under the Receiver Operating Characteristic Curve)**[14], [15]

The AUROC metric evaluates a model's ability to distinguish between in-distribution (ID) and out-of-distribution (OOD) samples across all possible classification thresholds. It quantifies the probability that a randomly chosen ID sample will be assigned a higher confidence score than an OOD sample. A higher AUROC (closer to 1.0) indicates better separation of our model between ID and OOD data.

### 4.2. Experimental Results

For a fair comparison, all hyperparameters identical to those in CoOp are kept unchanged. Specifically, we select 4 learnable prompts initialized with "a photo of a". We use SGD as the optimizer, with the learning rate set to 1e-4 and reduced using cosine annealing. Additionally, the first epoch includes a warm-up phase with the learning rate set to 1e-5. In Equation 3-9, we set α to 0.1, β to 0.01, and γ to 0.001. The reported results are obtained after testing following 5 epochs of training. For fairness, the scoring function used in all tests is the maximum softmax value with temperature scaling.

To demonstrate the superiority of the model, we selected two comparative methods. The first is zero-shot CLIP, representing the native classification capability of CLIP. The second is CoOp, representing the prompt learning method, used to test whether the prompt learning approach remains sufficiently robust when the training and test set distributions differ significantly.

We alternately treated each of the 4 domains as the source and target domains, conducting a total of 12 sets of experiments. Each set of experiments recorded three metrics: AUROC, FPR@TPR95, and CCR@FPR10. The results are shown in the table below.

In each experiment, the left side of the arrow represents the source domain, and the right side represents the target domain. "Pr" denotes "Product," "Rw" denotes "Real-World," "Cl" denotes "Clipart", "Ar" denotes "Art." Among the metrics: CCR10 represents CCR@FPR10, where a higher value indicates better performance. FPR95 represents FPR@TPR95, where a lower value indicates better performance. AUROC is better when higher. The best metric in each column is highlighted in bold black.

Table 1: The results whose Source Domain is Product.

| Method | Pr→Rw | | | Pr→Cl | | | Pr→Ar | | |
|---|---|---|---|---|---|---|---|---|---|
| | Acc10 | FPR95 | AUROC | Acc10 | FPR95 | AUROC | Acc10 | FPR95 | AUROC |
| CLIP | 76.97 | **31.05** | 92.14 | 51.00 | **79.63** | 78.56 | 63.19 | 59.16 | 85.84 |
| CoOp | 70.40 | 48.25 | 88.58 | 42.88 | 83.54 | 76.39 | 47.48 | 67.76 | 80.76 |
| Ours | **82.43** | 31.72 | **92.91** | **57.57** | 80.75 | **79.56** | **67.23** | **55.05** | **87.10** |

Table 2: The results whose Source Domain is Art

| Method | Ar→Pr | | | Ar→Rw | | | Ar→Cl | | |
|---|---|---|---|---|---|---|---|---|---|
| | Acc10 | FPR95 | AUROC | Acc10 | FPR95 | AUROC | Acc10 | FPR95 | AUROC |
| CLIP | 76.14 | **42.86** | 91.34 | 76.97 | **31.05** | 92.14 | 51.00 | **79.63** | 78.56 |
| CoOp | 74.97 | 49.34 | 90.05 | 76.91 | 42.64 | 90.71 | 50.82 | 82.87 | 77.86 |
| Ours | **77.32** | 43.65 | **91.51** | **79.56** | 32.97 | **92.38** | **54.16** | 80.08 | **79.44** |

Table 3: The results whose Source Domain is Real-World

| Method | Rw→Ar | | | Rw→Pr | | | Rw→Cl | | |
|---|---|---|---|---|---|---|---|---|---|
| | Acc10 | FPR95 | AUROC | Acc10 | FPR95 | AUROC | Acc10 | FPR95 | AUROC |
| CLIP | 63.19 | 59.16 | 85.84 | 76.14 | 42.86 | 91.34 | 51.00 | **79.63** | 78.56 |
| CoOp | 58.50 | 58.27 | 84.62 | **80.79** | 40.94 | **92.29** | 48.91 | 82.61 | **77.59** |
| Ours | **65.85** | **52.89** | **87.15** | 78.27 | **42.07** | 91.96 | **51.24** | 82.27 | 78.48 |

Table 4: The results whose Source Domain is Clipart

| Method | Cl→Rw | | | Cl→Ar | | | Cl→Pr | | |
|---|---|---|---|---|---|---|---|---|---|
| | Acc10 | FPR95 | AUROC | Acc10 | FPR95 | AUROC | Acc10 | FPR95 | AUROC |
| CLIP | 76.97 | **31.05** | 92.14 | 63.19 | 59.16 | 85.84 | 76.14 | 42.86 | 91.34 |
| CoOp | 78.85 | 44.44 | 90.84 | **66.03** | 61.48 | 85.34 | **78.27** | 43.24 | 91.76 |
| Ours | **79.89** | 32.03 | **92.16** | 62.54 | **55.35** | **86.38** | 77.94 | **39.32** | **91.87** |

As can be observed, when there is a significant distribution gap between the target and source domains, the performance of CoOp—a fine-tuning method based on the independent and identically distributed (i.i.d.) assumption between training and test sets—deteriorates substantially. This manifests as degraded quality of the learned continuous prompts, not only weakening the model's recognition capability for in-distribution samples but also resulting in worse FPR95 and AUROC metrics compared to zero-shot CLIP. This indicates that CoOp struggles to handle distribution shifts effectively.

Ablation results are as follows.

Table 5: Ablation studies. The results are averaged over 12 tasks with different random seeds.

| Method | Avg | | |
|---|---|---|---|
| | Acc10 | FPR95 | AUROC |
| CLIP | 66.82 | 53.17 | 86.97 |
| "+CE" | **68.15** | **51.67** | **87.47** |
| "+KL" | 66.51 | 52.98 | 86.05 |

As shown in the table, when the model is only CLIP, the result appear common performance. However, as "+CE" is added, the performance of model is improved. "+CE" means we use cross entropy as our loss function which employs pseudo labels to approach labels of identified shared class in the target domain. Then, "+KL" means the Kullback-Leibler divergence between uniform distribution and the distribution of pseudo labels which are identified unknown classes by our proposed model. The performance is still higher than the baseline.

## 5. Conclusion

From our survey, we propose two ideas: 1) Prompt-driven cross-domain alignment, 2) Gradient-aware open-set separation to break through the limitation of current approaches. However, we find that the results of experiment is opposite of GradNorm. Then, we infer the back propagation of our model and finally find that the reason. Eventually, we continue the experiment and get the understanding performance of the model.